%% file: main.tex
\documentclass[10pt,twocolumn,letterpaper]{article}

\usepackage{cvpr}              

\input{preamble}

\definecolor{cvprblue}{rgb}{0.21,0.49,0.74}
\usepackage[pagebackref,breaklinks,colorlinks,allcolors=cvprblue]{hyperref}

\def\etal{\textit{et~al}.\ }

\def\paperID{39651} 
\def\confName{CVPR}
\def\confYear{2026}

\title{Linear Image Generation by Synthesizing Exposure Brackets}

\author{
Yuekun Dai$^{1}$ $\,\,\,\,$ Zhoutong Zhang$^{2}$ $\,\,\,\,$ Shangchen Zhou$^{1}$ $\,\,\, $ Nanxuan Zhao$^{3}$ \\
$^{1}$S-Lab, Nanyang Technological University \quad $^{2}$Adobe NextCam \quad $^{3}$Adobe Research \\
\texttt{\small \{ydai005, s200094\}@ntu.edu.sg, \{zhoutongz, nanxuanz\}@adobe.com }\\ \vspace{-6mm}
{\tt\small \url{https://ykdai.github.io/projects/raw_gen}}
}

\begin{document}
\twocolumn[{%
\renewcommand\twocolumn[1][]{#1}%
\maketitle
\vspace{-5mm}
\begin{center}
    \centering
    \captionsetup{type=figure}
    \includegraphics[width=1.0\textwidth]{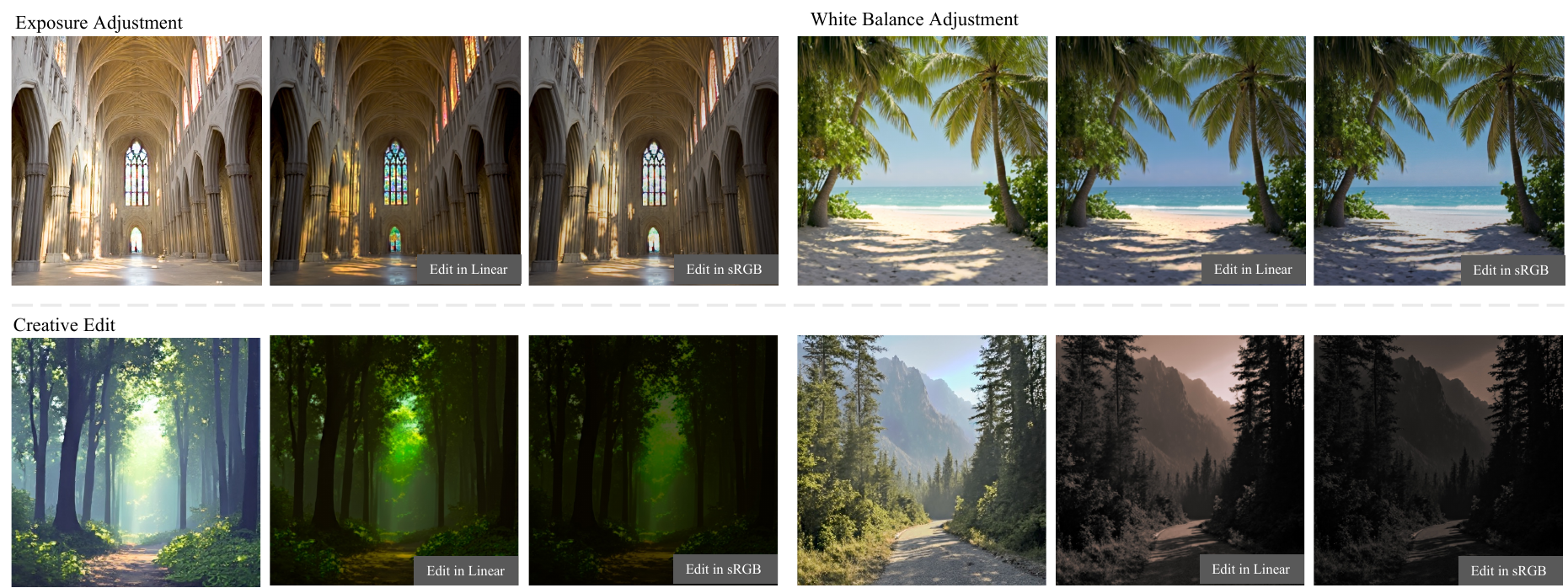}
    \vspace{-5.5mm}
    \captionof{figure}{Linear image provides more room for the user to edit compared to sRGB images. Here we show the difference between linear and sRGB images for post editing. All four images are generated from our model, and the same edits are applied in both linear space and sRGB to showcase the difference. Linear images provide more details in highlights, are more accurate for white balance adjustments and support more dramatic creative edits.
    }
    \label{fig:production_pipeline}
\end{center}%
}]
\maketitle
{
  \renewcommand{\thefootnote}%
    {\fnsymbol{footnote}}
  \footnotetext[1]{Work done during internship at Adobe.}
}
\input{sec/0_abstract}

\input{sec/1_intro}
\input{sec/2_related_work}
\input{sec/3_dataset}
\input{sec/4_method}
\input{sec/5_experiments}
\input{sec/6_conclusion}

\clearpage
{
    \small
    \bibliographystyle{ieeenat_fullname}
    \bibliography{main}
}

\input{sec/X_suppl}
\end{document}

%% file: preamble.tex
\usepackage{algorithm}
\usepackage{algorithmic}
\usepackage{multirow}

%% file: sec/0_abstract.tex
\begin{abstract}
The life of a photo begins with photons striking the sensor, whose signals are passed through a sophisticated image signal processing (ISP) pipeline to produce a display-referred image.
However, such images are no longer faithful to the incident light, being compressed in dynamic range and stylized by subjective preferences.
In contrast, RAW images record direct sensor signals before non-linear tone mapping. After camera response curve correction and demosaicing, they can be converted into linear images, which are scene-referred representations that directly reflect true irradiance and are invariant to sensor-specific factors.
Since image sensors have better dynamic range and bit depth, linear images contain richer information than display-referred ones, leaving users more room for editing during post-processing.
Despite this advantage, current generative models mainly synthesize display-referred images, which inherently limits downstream editing.
In this paper, we address the task of \textbf{text-to-linear-image generation}: synthesizing a high-quality, scene-referred linear image that preserves full dynamic range, conditioned on a text prompt, for professional post-processing.
Generating linear images is challenging, as pre-trained VAEs in latent diffusion models struggle to simultaneously preserve extreme highlights and shadows due to the higher dynamic range and bit depth.
To this end, we represent a linear image as a sequence of exposure brackets, each capturing a specific portion of the dynamic range, and propose a DiT-based flow-matching architecture for text-conditioned exposure bracket generation.
We further demonstrate downstream applications including text-guided linear image editing and structure-conditioned generation via ControlNet.

\end{abstract}

%% file: sec/1_intro.tex
\section{Introduction}
\label{sec:intro}
Most of the images we encounter every day are stylized renditions of the real world: they are the results of complex imaging pipelines where every pixel encodes display colors rather than how bright the underlying scene truly is. These images are typically referred to as display-referred images. In contrast, linear images are scene-referred representations that record the scene irradiance at each pixel location on the imaging plane, prior to any nonlinear tone mapping or compression.
A linear image can be derived from a RAW capture by correcting the camera response curve, performing demosaicing, and converting to a camera-independent color space. Through these operations, linear images become faithful and sensor-agnostic descriptions of physical radiance, providing a robust foundation for post-processing, relighting as shown in Fig.~\ref{fig:production_pipeline}, as well as benefiting some computational photography applications~\cite{kee2025removing,huang2022towards,cai2025parametric,wang2025flash}.

\begin{figure}[t]
  \centering
  \includegraphics[width=0.9\linewidth]{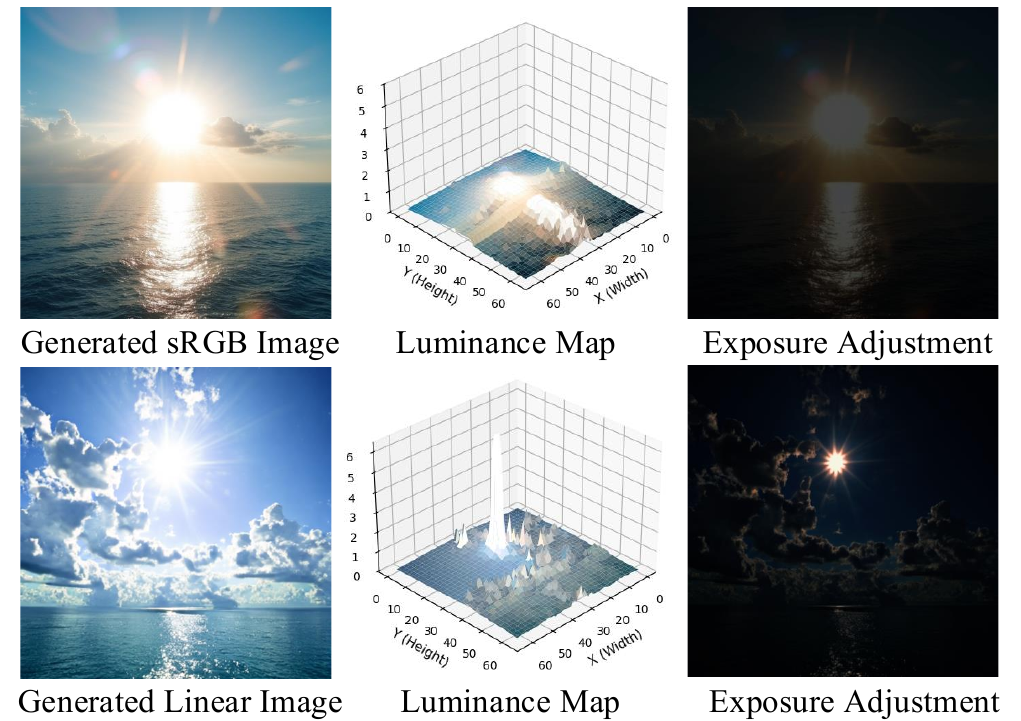}
  \vspace{-1mm}
    \caption{Linear images are scene-referred representations that directly reflect the irradiance at the imaging plane, without the nonlinear tone mapping or compression applied in sRGB or HDR images. Their high bit depth and dynamic range support flexible exposure adjustments without losing highlight details.}
    \vspace{-2mm}
  \label{fig:linear}
\end{figure}

Modern image generative models~\cite{rombach2022high,esser2024scaling,chen2024pixart} are trained almost exclusively on display-referred datasets like LAION-5B~\cite{schuhmann2022laion}. 
Consequently, they are capable of producing aesthetically pleasing yet tone-mapped results whose brightness and contrast are constrained by display dynamic range.
When generating scenes containing both bright highlights and deep shadows, these models tend to produce flattened tone-mapped images with limited post-editing flexibility. By contrast, linear images preserve the full dynamic range of the scene, offering photographers and downstream systems significantly greater room for exposure, white-balance, and tone adjustment without introducing highlight clipping as shown in Fig.~\ref{fig:linear}.
Despite the advantages of linear images, research on their generative modeling remains limited.
Ideally, a user should be able to generate a linear image directly from a text description, obtaining a scene-referred result ready for professional post-processing without any lossy conversion.
At the same time, AI-based image editing tools are widely used, but they are designed for display-referred images and cannot directly process linear images. As a result, users must convert linear images to sRGB before editing, discarding dynamic range information that cannot be recovered after the fact. This gap highlights the need for \emph{text-to-linear-image} generative models and downstream editing tools that natively support linear image formats.

\begin{figure}[t]
  \centering
  \includegraphics[width=1.0\linewidth]{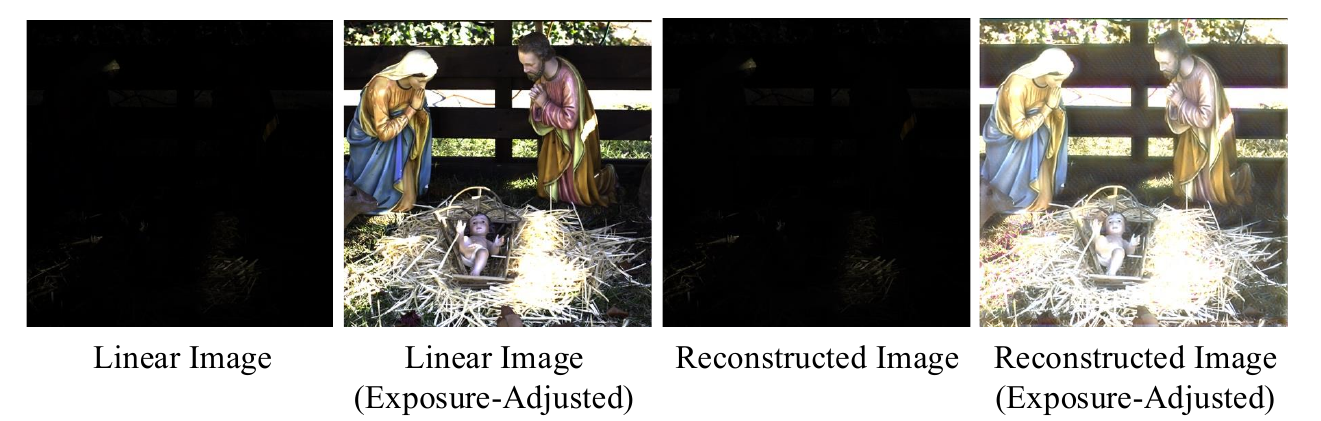}
  \vspace{-6mm}
    \caption{Visual comparison of brightened VAE reconstructed image and exposure-adjusted ground truth. The VAE fails to preserve details in very dark region in linear images, causing severe information loss after the encode-decode process. We use the VAE's float32 precision to mitigate quantization artifacts in this case.}
    \vspace{-2mm}
  \label{fig:reconstruction}
\end{figure}

Building a generative model for linear images is inherently challenging.
First, data scarcity poses a major bottleneck: scene-referred images typically remain private to photographers, while only their tone-mapped versions are publicly shared. This makes it impractical to train large-scale diffusion models purely from linear data.
Second, the high dynamic range and bit depth of linear images exceed the representational capacity of the pre-trained variational autoencoder (VAE) used in latent diffusion models. VAEs are trained to reconstruct display-referred images within limited value ranges; when applied to linear data, they struggle to simultaneously preserve details in both highlights and shadows, as shown in Fig.~\ref{fig:reconstruction}. The result resembles an image captured by a sensor with insufficient dynamic range, where regions of extreme brightness or darkness are clipped or compressed.
This analogy motivates our solution: inspired by exposure bracketing in photography, we represent a linear image as a sequence of exposure-specific sub-images, each capturing a distinct portion of the overall dynamic range. By reconstructing these exposure brackets instead of a single high-dynamic-range frame, we circumvent the limitations of existing VAEs and enable reliable synthesis of high-bit-depth content.
Specifically, we develop a flow-matching-based generative framework that synthesizes multiple exposure brackets and fuses them into a unified linear image. The framework builds upon Flux~\cite{flux2024}, enhanced with exposure modulation self-attention and LoRA.
We further introduce a radiance scale-token denoising mechanism for joint radiance scale-image generation, enabling explicit prediction of the scene radiance scale. Our proposed exposure modulation self-attention also ensures consistency across exposure brackets, allowing the model to generate well-aligned exposure brackets and scene radiance scale jointly. 

Extensive experiments demonstrate that our method produces realistic, physically accurate linear images with rich dynamic range, and enables training-free linear image editing as well as ControlNet-guided conditional generation.
Our main contributions can be summarized as follows:
(1) A text-conditioned flow-matching framework for high-bit-depth linear image generation, employing multi-exposure synthesis to expand dynamic range beyond VAE limits.
(2) A radiance-scale token denoising mechanism for joint prediction of radiance scale and image content, improving scene radiance reconstruction.
(3) Integration of exposure modulation self-attention and LoRA fine-tuning within a DiT backbone, achieving efficient and high-fidelity synthesis.
(4) Comprehensive evaluations and applications demonstrating realistic text-to-linear-image generation and downstream tasks including linear image editing.

%% file: sec/2_related_work.tex
\section{Related Work}
\label{sec:related_work}
\noindent{\bf HDR Image Reconstruction and Generation.}
Similar to our method, HDR image reconstruction and generation also aim to recover high dynamic range content.
Most existing HDR generation methods~\cite{eilertsen2017hdr, chen2022text2light, wang2022stylelight, wang2025lediff, guan2025hdr} adopt a two-stage pipeline, where an LDR image is first synthesized and then converted to HDR via an inverse tone-mapping (ITM) module.
Eilertsen~\etal~\cite{eilertsen2017hdr} introduce deep learning to the ITM task by employing convolutional neural networks (CNNs) to enhance highlight details in HDR reconstruction.
Following this paradigm, Text2Light~\cite{chen2022text2light} and StyleLight~\cite{wang2022stylelight} focus on HDR panorama generation by first estimating the LDR panorama and subsequently training an ITM network for HDR estimation.
LEDiff~\cite{wang2025lediff} adopts a fused latent diffusion framework to progressively infer shadow and highlight regions, thereby achieving a more accurate and continuous HDR formation process.
Guan~\etal~\cite{guan2025hdr} further propose a two-stage diffusion-based pipeline that first synthesizes standard dynamic range (SDR) images using a pretrained diffusion model and then estimates a gain map via adapted stable diffusion.
In contrast to the aforementioned two-stage approaches, several recent methods~\cite{wang2023glowgan, bemana2025bracket} focus on direct HDR generation without relying on intermediate LDR representations.
GlowGAN~\cite{wang2023glowgan} leverages Gaussian exposure modeling to capture HDR-LDR relationships, enabling unsupervised HDR synthesis and improved reconstruction of over-exposed areas using pretrained generative priors.
Bracket Diffusion~\cite{bemana2025bracket} leverages Denoising Diffusion Probabilistic Models (DDPMs)~\cite{ho2020ddpm} to directly simulate multi-exposure bracketed imaging for HDR synthesis through test-time optimization.
Among these, only GlowGAN and Bracket Diffusion are capable of direct HDR generation.
However, GlowGAN is restricted to specific image categories, while Bracket Diffusion takes several minutes to generate a single 256$\times$256 HDR image.
Since HDR images are typically produced through tone-mapping operations, their generation priors are difficult to leverage for downstream linear image tasks, and these methods do not support HDR-related editing, limiting their applicability in content creation workflows.

\noindent{\bf RAW/Linear Image Reconstruction.}
Although no prior work has directly addressed text-to-RAW/Linear image generation, numerous studies have focused on sRGB-to-RAW reconstruction.
The same RAW image can be rendered into different sRGB images through various in-camera Image Signal Processing (ISP) pipelines.
Thus, sRGB-to-RAW is an ill-posed problem without any RAW-image-prior, most RAW reconstruction methods assume the RAW to sRGB mapping follows a certain pattern to avoid this ill-posedness.
UPI~\cite{brooks2019unprocessing} assumes that the ISP applies global tone mapping and introduces a method to synthesize realistic RAW data by inverting the standard camera ISP pipeline, enabling effective training of denoising models on unpaired data.
CycleISP~\cite{zamir2020cycleisp}, InvISP~\cite{xing2021invertible} and ReverseISP~\cite{conde2022reversed} propose paired learning-based sRGB-to-RAW and RAW-to-sRGB networks, which assume the generated sRGB images are produced by a specific invertible ISP pipeline.
RAW-Diffusion~\cite{reinders2025rawdiffusion} uses DDPM~\cite{ho2020ddpm} for sRGB-to-RAW image generation, also following the global tone mapping ISP.
Other methods take additional metadata~\cite{yuan2011high,nguyen2016raw,punnappurath2021spatially,nam2022learning,wang2023raw} or reference image~\cite{otsuka2023self} for sRGB to RAW image reconstruction.
The requirements of the metadata and specific ISP can all be attributed to the lack of a RAW image generation prior.
Therefore, in this paper, we aim to address this problem through a text-to-linear-image generation framework to establish this kind of prior.

\noindent{\bf RAW Image Dataset.}
Unlike sRGB images, RAW data are less publicly available due to their large storage requirements and the need for specialized camera capture settings.
The Adobe FiveK dataset~\cite{fivek} is one of the most widely used benchmarks for RAW image processing, containing 5,000 RAW-sRGB image pairs retouched by professional photographers.
Another commonly used dataset is RAISE~\cite{dang2015raise}, which includes 8,156 high-resolution RAW photographs captured under diverse lighting conditions.
In this work, we collect additional RAW images and use them together with RAISE~\cite{dang2015raise} as the training data, while adopting Adobe FiveK~\cite{fivek} as the evaluation set for our experiments.

%% file: sec/3_dataset.tex
\section{Data Collection and Processing}
\label{sec:dataset}

\begin{figure*}[ht]
    \centering
    \includegraphics[width=\textwidth]{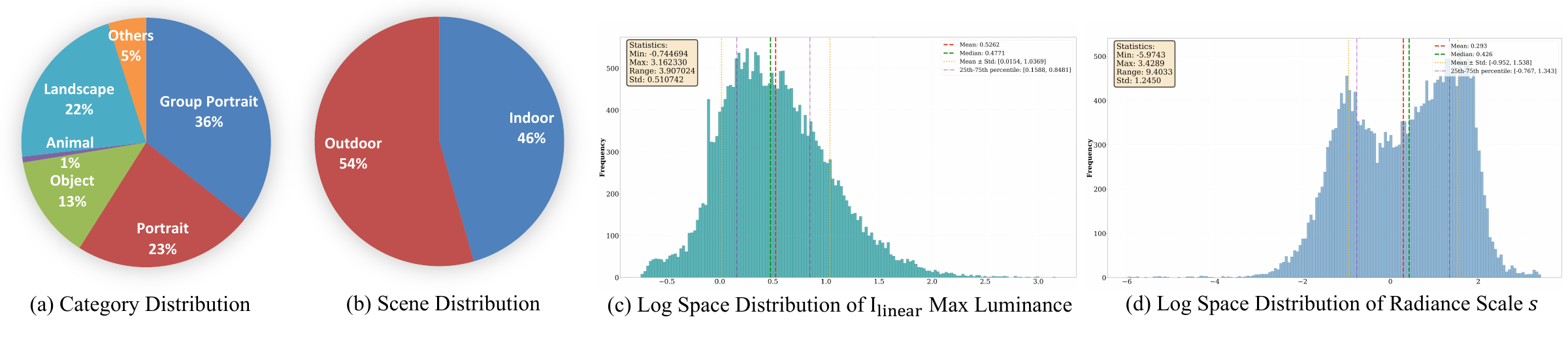}
    \vspace{-7mm}
    \caption{Overview of our dataset statistics. We show the overall composition of our dataset in terms of content categories and scene types, along with the distributions (in $\log_{10}$ space) of the radiance scale $s$ and linear image's luminance values. Larger radiance scale means the image is brighter in the physical world. These distributions highlight the large dynamic-range variation of the linear image.}
    \vspace{-2mm}
    \label{fig:data_pipeline}
\end{figure*}

\subsection{Data Preprocessing}
Acquiring a large number of linear images is extremely challenging in practice. Moreover, most public HDR datasets are either panoramic (thus focusing almost exclusively on large-scale scene content) or do not provide true linear images, making them unsuitable for our purposes. Therefore, we primarily use RAW image datasets as the basis for training. As discussed in the previous section, we use RAISE~\cite{dang2015raise} and our own collected RAW images as the training set, and adopt the MIT-Adobe FiveK~\cite{fivek} as our test set. To ensure high visual quality, we filter out images with aesthetic scores below 4.5, ultimately retaining 25k images in our training set.

We develop a preprocessing pipeline to convert camera-specific RAW data into camera-independent, scene-referred linear images. Our pipeline consists of four main stages: (1) demosaicing to convert Bayer pattern data to full RGB; (2) applying color correction matrices (CCM) and lookup tables (LUT) to transform sensor signals into camera-independent RGB space; (3) white balance adjustment to standardize color temperature to 5000K; and (4) converting RGB to CIE-XYZ for denoising, then back to linear RGB as linear sensor signals $\mathbf{I}_s$.

Given the processed sensor signal $\mathbf{I}_s$, we recover scene radiance via the standard exposure inversion:
\begin{equation}
\mathbf{L}
= \mathbf{I}_s \cdot \frac{F^2}{t \cdot ISO}\cdot 2^{-EV},
\end{equation}
where $t$, $ISO$, and $F$ denote exposure time, sensor gain, and aperture $F$-number, and $EV$ is the exposure-compensation setting in stops.

To stabilize training and avoid extremely wide range of radiance values, we normalize $\mathbf{L}$ using a radiance scale $s$ to obtain the normalized linear image $\mathbf{I}_{\text{linear}}$.  
Instead of enforcing a fixed median value, we compute $s$ from two robust percentile statistics of the radiance distribution of $\mathbf{L}$.  
Specifically, we estimate a mid-level radiance and a highlight radiance as:
\begin{equation}
m = \operatorname{Percentile}_{0.5}(\mathbf{L}), \qquad
h = \operatorname{Percentile}_{0.9}(\mathbf{L}),
\end{equation}
and derive two candidate radiance scales:
\begin{equation}
s_{\text{med}} = \frac{m}{0.18}, 
\qquad
s_{\text{hi}} = \frac{h}{0.8}.
\end{equation}

To address cases where large dark backgrounds might cause subject regions to become abnormally bright after normalization, we incorporate the highlight-based radiance scale $s_{\text{hi}}$ to help constrain subject exposure. We use the maximum of the two candidate radiance scales as:
\begin{equation}
s = \max(s_{\text{med}},\, s_{\text{hi}}).
\end{equation}

Finally, the normalized linear image used for training is obtained as:
\begin{equation}
\mathbf{I}_{\text{linear}} = \frac{\mathbf{L}}{s}.
\label{eq:linear_image}
\end{equation}
This percentile-based normalization effectively controls overall scene radiance while preventing highlight saturation, producing linear images which are suitable for training. 
The distribution of radiance scale $s$ and $\mathbf{I}_{\text{linear}}$ is shown in Fig.~\ref{fig:data_pipeline}, together with the category and scene compositions of our dataset.

\subsection{Image Captioning}
To create text labels, we use the Qwen2.5-VL 7B~\cite{qwen2.5-VL} with the instruction ``Describe the content and details of the image in a direct, concise way, without introductory phrases.'' to generate captions for the EV0 (base exposure) images in our dataset. The instruction style is chosen to match the caption distribution of the base Flux model's pretraining data. The generated captions serve as the text conditioning signal during training, enabling text-to-linear-image generation. 

%% file: sec/4_method.tex
\section{Proposed Method}
\label{sec:method}

\begin{figure*}[ht]
    \centering
    \includegraphics[width=1.0\textwidth]{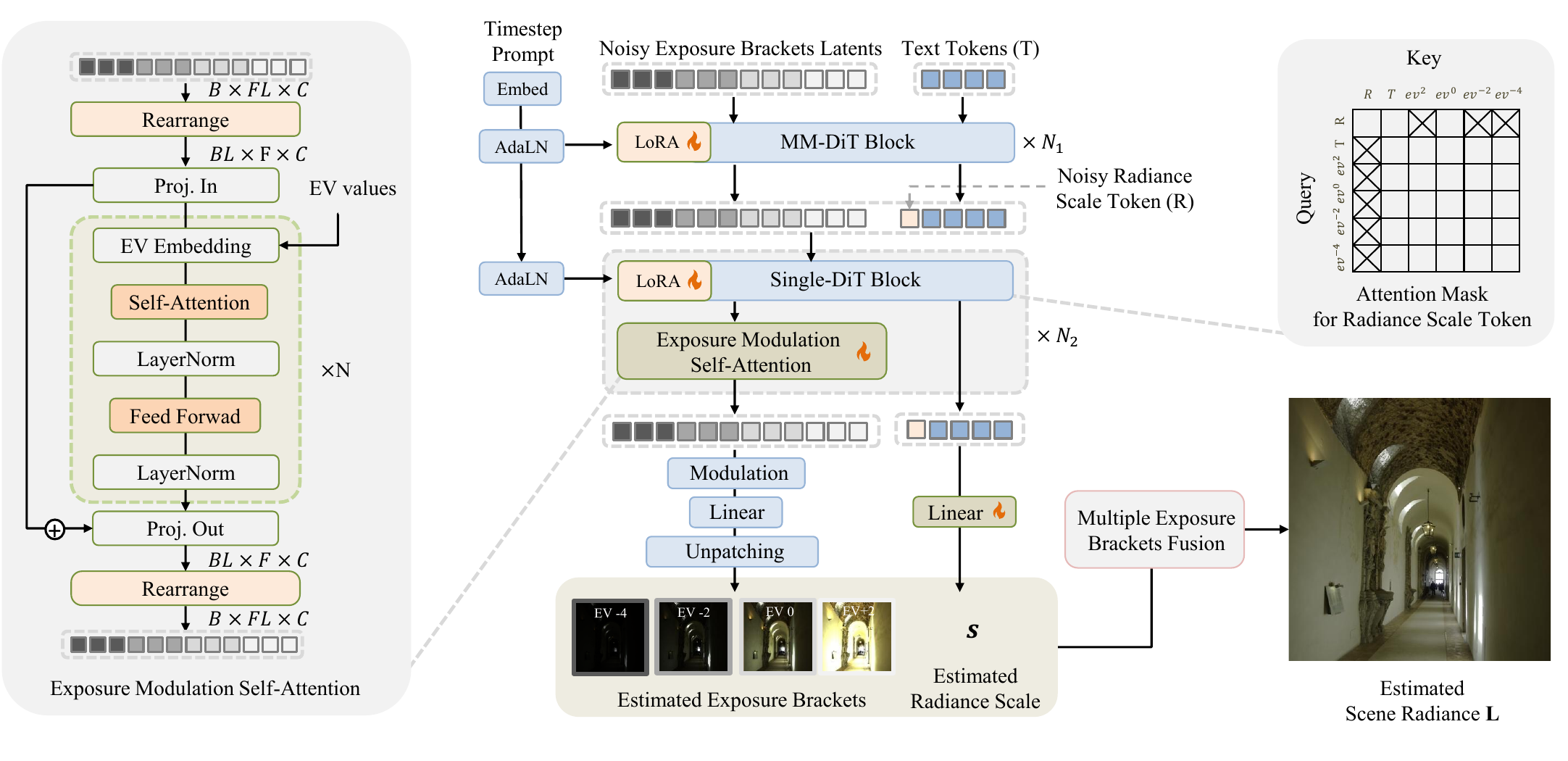}
    \vspace{-10mm}
    \caption{Overview of the proposed framework. The model takes as input the concatenated noise of $K{=}4$ exposure brackets, text prompt, and a noisy radiance-scale token $R$. These are processed through $N_1$ MM-DiT blocks (with LoRA), followed by $N_2$ Single-DiT blocks (with LoRA) that incorporate our Exposure Modulation Self-Attention for exposure-aware feature modulation. The radiance-scale token attends only to text tokens and the EV0 bracket (as shown in the attention mask), and is projected via a linear layer to produce the estimated radiance scale $s$. The denoised bracket tokens are decoded into $K$ exposure brackets, which are then fused by our Multiple Exposure Brackets Fusion module to yield the final scene-referred linear image $\mathbf{L}$.}
    \label{fig:framework}
    \vspace{-5mm}
\end{figure*}

\subsection{Problem Formulation}
Our method aims to jointly predict a radiance scale $s$ and its corresponding scene-referred linear image $\mathbf{I}_{\text{linear}} \in \mathbb{R}^{H\times W\times 3}$, both conditioned on input text prompts. 
Due to the high bit depth and extensive dynamic range of linear images, it is difficult to directly reconstruct linear images using a VAE. To address this challenge, we decompose the linear image into a sequence of exposure brackets, each capturing the scene at a different exposure level.
Given a list of exposure values $\mathrm{EV} = [ev_1, ev_2, ..., ev_K]$, each bracket image $\mathbf{I}_k$ is constructed from the normalized linear image:
\begin{equation}
\mathbf{I}_k = \operatorname{clip}(\mathbf{I}_{\text{linear}} \cdot 2^{ev_k},\; 0, 1),
\end{equation}
where $\operatorname{clip}(\cdot,\;0,1)$ restricts the values to the range $[0,1]$. In this work, we adopt $\mathrm{EV} = [-4, -2, 0, 2]$ as our set of exposure values, resulting in $K=4$ bracket images derived from each linear image.
To enable independent encoding through the shared VAE encoder without modification, the $K$ exposure bracket images $\mathbf{I}_k$ are concatenated along the batch dimension and passed through a shared VAE encoder. This produces $K$ latent sequences $\mathbf{z}_k \in \mathbb{R}^{L \times C}$, where $L$ is the sequence length and $C$ is the embedding dimension. These latent sequences are then concatenated along the sequence dimension to form a combined latent representation $\mathbf{z}_{\text{all}} \in \mathbb{R}^{K L \times C}$, which aggregates information from all exposure brackets.

In our approach, we begin by initializing two sets of Gaussian noise: one latent tensor $\mathbf{z}_t \in \mathbb{R}^{KL \times C}$ for the exposure brackets and another noise token of shape $(1 \times C)$ for the radiance scale. Through flow-matching denoising, these noisy representations are transformed into the corresponding exposure bracket tokens and a radiance scale token. The exposure bracket tokens are subsequently decoded by the VAE decoder to generate multi-exposure bracket images, while the radiance scale token is projected back to a scalar via a linear layer for radiance scale estimation. This procedure enables the joint generation of both exposure brackets and the global radiance scale.

\subsection{Architecture Design}

\noindent{\bf LoRA Finetuning.}
LoRA fine-tuning adapts the pre-trained Flux backbone to the linear image domain with minimal parameter overhead. It prevents the Single-DiT backbone from destabilizing under the rapidly changing exposure distributions across different bracket frames.

\noindent{\bf Exposure Modulation Self-Attention.}
As shown in Fig.~\ref{fig:framework}(b), we introduce Exposure Modulation Self-Attention, which jointly attends across all exposure brackets. This mechanism allows the model to flexibly adjust luminance for each bracket while preserving structural alignment and detail consistency across exposures.

\begin{figure*}[ht]
    \centering
    \vspace{-3mm}
    \includegraphics[width=\textwidth]{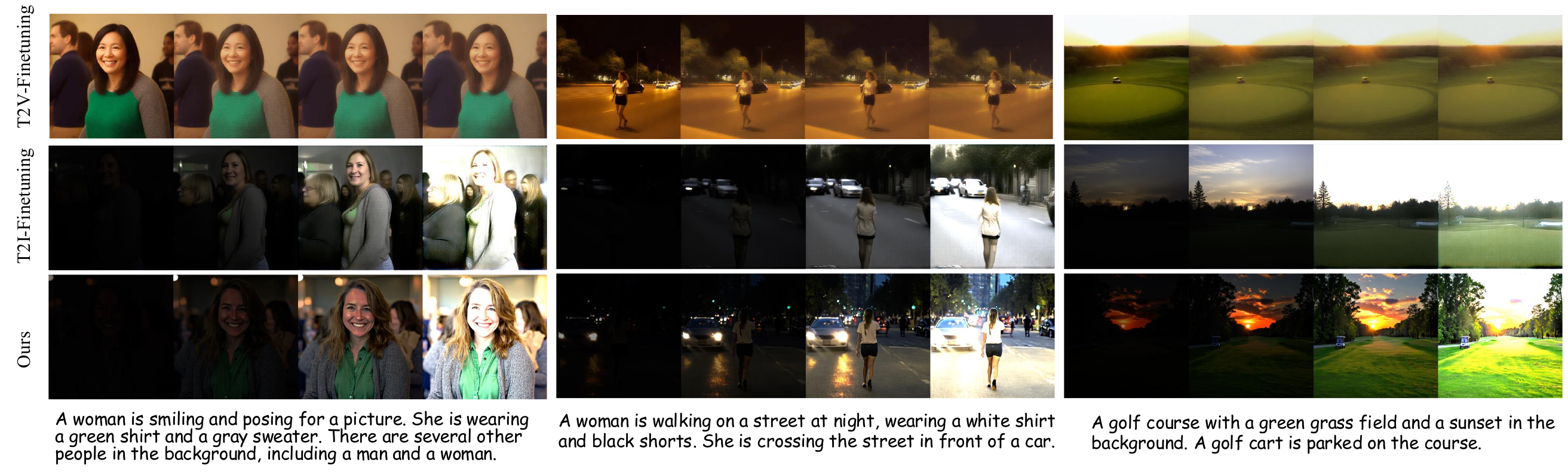}
    \vspace{-7mm}
    \caption{Comparison of our proposed method with LoRA finetuning on Flux and Wan~\cite{wan2025wan}. Unlike other baseline methods, our approach effectively preserves shadow details even in over-exposed situations.}
    \vspace{-4mm}
    \label{fig:comparison_methods}
\end{figure*}

\noindent{\bf 3D-RoPE for Sequence Disentanglement.}
We use 3D Rotary Positional Embedding (3D-RoPE) to encode both spatial position and bracket identity into each token, helping the model distinguish and correctly process tokens from different exposure brackets.
In the standard Flux formulation, each image token is assigned a 2D positional coordinate $(i, j)$ corresponding to its spatial location in the image grid.
To extend this to multi-bracket generation, we augment the positional encoding to a 3D tuple $(\mathrm{index},\, i,\, j)$, where $i$ and $j$ retain their original spatial meaning, and $\mathrm{index}$ is set to the bracket index $k \in \{0, 1, \ldots, K{-}1\}$ of the exposure frame to which the token belongs.
This design allows the model to unambiguously identify which exposure bracket each token belongs to while preserving intra-bracket spatial structure, enabling effective disentanglement of luminance levels across the jointly attended sequence.
\subsection{Radiance Scale Tokenization and Prediction}
For quantization and prediction, we use $s_l = \log_{10}(s)$, the log-transformed radiance scale.
We discretize the ground-truth log-radiance $s_l$ into $20$ uniform bins over the interval $[-6,4]$.  
Each $s_l$ value is mapped to a one-hot discrete code $\mathbf{s}_{d}\in\{0,1\}^{20}$ indicating its quantized bin.  
This discrete code is then embedded into the diffusion token space through a shared linear projection:
\begin{equation}
\mathbf{t}_{s} = W\,\mathbf{s}_{d}, \qquad W\in\mathbb{R}^{20\times d},
\end{equation}
which is then jointly updated with image tokens during self-attention.  
At inference time, the updated radiance token $\hat{\mathbf{t}}_{s}$ is projected back to the bin space:
\begin{equation}
\hat{\mathbf{s}}_{d}= \mathrm{softmax}(\hat{\mathbf{t}}_{s} W^\top),
\end{equation}
and the predicted log-radiance $\hat{s}_l$ is obtained as the expectation over the $20$ bin centers $\mu_i$:
\begin{equation}
\hat{s}_l = \sum_{i=1}^{20} \hat{\mathbf{s}}_{d}[i]\;\mu_i.
\end{equation}
If recovering the predicted radiance scale $s$ is needed, it can be computed by $s = 10^{\hat{s}_l}$. 

To ensure robust radiance scale prediction, we carefully design an attention mask for the radiance scale token, as illustrated in Fig.~\ref{fig:framework}. Specifically, the radiance scale token is only allowed to attend to the text tokens and the tokens corresponding to the EV0 (base exposure) bracket. This allows the model to derive a global scene radiance primarily from the reference exposure, mitigating the risk of being influenced by the over- or under-exposed bracket images. Conversely, the other image and text tokens do not attend to the radiance scale token, aiming to maintain the fidelity of generated images. Additionally, the radiance scale token does not undergo 3D-RoPE positional embedding, as it does not represent a spatial location or exposure index.To improve the accuracy of radiance scale estimation, we add a pooling layer over the text tokens, and design a linear decoder that estimates the radiance scale from the pooled text token and the radiance scale token which is initialized using radiance scale's decoder.

\subsection{Training Objective}
Our model is trained under a flow-matching formulation.  
Let $\mathbf{z}_t$ denote the latent at continuous time $t\in[0,1]$, and let $\phi_t(\mathbf{z}_0,\mathbf{z}_1)$ be the ground-truth probability flow that maps clean latents $\mathbf{z}_0$ to noise $\mathbf{z}_1$.  
Flow matching constructs a velocity field $u_t(\mathbf{z}_t)$ satisfying the probability flow ODE
\begin{equation}
\frac{d\mathbf{z}_t}{dt} = u_t(\mathbf{z}_t),
\end{equation}
and the model learns to approximate this field via  
\begin{equation}
\mathcal{L}_{\text{img}}
= \mathbb{E}_{t,\mathbf{z}_0,\mathbf{z}_1}
\left[
\left\|
u_t(\mathbf{z}_t)
-
u_\theta(\mathbf{z}_t, \mathbf{c}, t)
\right\|_2^2
\right].
\end{equation}

The radiance-scale token is supervised in the same velocity space: if $\mathbf{t}_{s,t}$ and $\hat{\mathbf{t}}_{s,t}$ denote the ground-truth and predicted radiance tokens following the probability flow, then
\begin{equation}
\mathcal{L}_{\text{rad}}
=
\left\|
u_t(\mathbf{t}_{s,t})
-
u_\theta(\hat{\mathbf{t}}_{s,t})
\right\|_2^2,
\end{equation}
where $u_t$ denotes analytically defined token-space velocity derived from the quantized radiance interpolation schedule.

To ensure physically consistent exposure ordering among the generated exposure brackets, we further impose a bracket-consistency loss computed in pixel space.  
After denoising, we obtain the predicted clean latents $\widehat{\mathbf{z}}_0$, decode them with the VAE, reshape them into $(B,F,C,H,W)$, and enforce that all frames align with the reference exposure (EV0).  
Let $\hat{\mathbf{I}}_k$ denote the predicted frame at exposure $ev_k$ and let $k_0$ be the index where $ev_k=0$, so that $\hat{\mathbf{I}}_{k_0}$ is the predicted EV=0 (base exposure) bracket serving as the reference.
We compute
\begin{equation}
\mathcal{L}_{\text{bracket}}
=
\sum_{k}
\left\|
\frac{\hat{\mathbf{I}}_{k}}{2^{ev_k}}
-
\hat{\mathbf{I}}_{k_0}
\right\|_1,
\end{equation}
which enforces the expected multiplicative radiance ratios between bracketed frames without relying on masking or thresholding.
The total loss is defined as:
\begin{equation}
\mathcal{L}
=
\mathcal{L}_{\text{img}}
+ \lambda_{\text{rad}} \mathcal{L}_{\text{rad}}
+ \lambda_{\text{bracket}} \mathcal{L}_{\text{bracket}},
\end{equation}
where $\lambda_{\text{rad}}$ and $\lambda_{\text{bracket}}$ are set to 1.0 and 0.5 respectively. 

\subsection{Multiple Exposure Brackets Fusion}
At inference time, we synthesize $K$ exposure-bracketed images $\hat{\mathbf{B}} = \{\hat{\mathbf{I}}_1, \ldots, \hat{\mathbf{I}}_K\}$ (from darkest to brightest) and subsequently fuse them to obtain the final linear image $\hat{\mathbf{I}}_{\text{linear}}$. 
For each pair of consecutive exposure images $\hat{\mathbf{I}}_{k}$ and $\hat{\mathbf{I}}_{k+1}$, we compute the mean intensity of each channel in non-saturated regions for each frame. 
We then compute a per-channel ratio vector $r_k \in \mathbb{R}^3$ by dividing the mean value of each channel in $\hat{\mathbf{I}}_{k+1}$ by the corresponding mean in $\hat{\mathbf{I}}_{k}$, where the mean is taken over valid (non-saturated) pixels.
This three-channel ratio ensures accurate alignment of brightness transitions across the exposure levels.

The fusion process proceeds in a hierarchical fashion, starting from the brightest bracket and iteratively incorporating information from darker brackets. At each step $k$ ($k = K-1, ..., 1$), a soft mask $\mathbf{M}_k \in [0, 1]^{H \times W \times 3}$ is constructed to delineate spatial regions where the current bracket provides unsaturated, high-fidelity information, with smoothing to avoid blending artifacts. The fusion at each stage is then performed as a weighted combination:
\begin{equation}
\hat{\mathbf{I}}_{\text{fused}} \leftarrow 
    \hat{\mathbf{I}}_{\text{fused}} \cdot (1 - \mathbf{M}_k)
    + \big(\hat{\mathbf{I}}_k \cdot r_k\big) \cdot \mathbf{M}_k,
\end{equation}
where $\hat{\mathbf{I}}_k \cdot r_k$ denotes per-channel multiplication to align the luminance of $\hat{\mathbf{I}}_k$ with the fused result. The fused image is initialized as $\hat{\mathbf{I}}_K$ and updated recursively. 
This method enables the reconstruction of a high-dynamic-range linear image from multiple exposure-biased generations, effectively recovering highlight and shadow details while ensuring radiometric consistency and smooth transitions.

%% file: sec/5_experiments.tex
\begin{figure}[t]
    \centering
    \includegraphics[width=1.0\columnwidth]{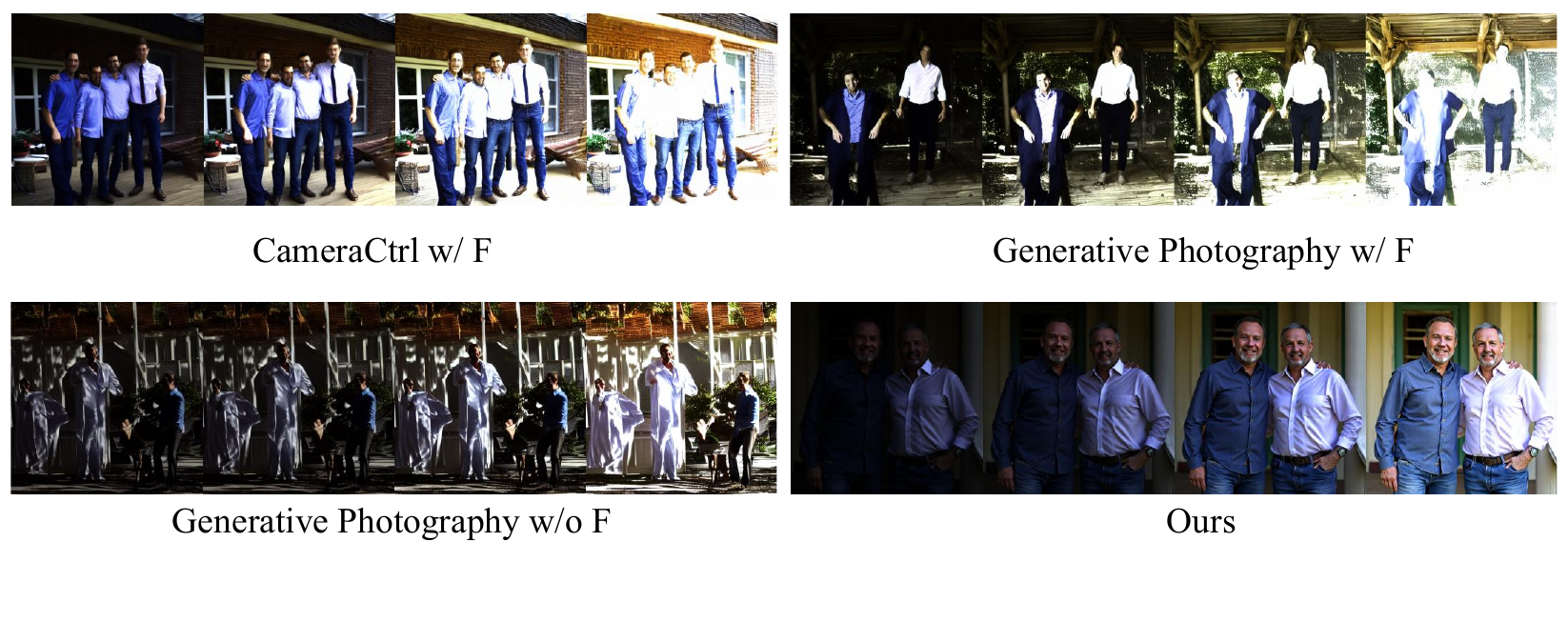}
    \vspace{-9mm}
    \caption{Visual comparison with CameraCtrl and Generative Photography. Our method produces better-quality exposure brackets with more consistent luminance transitions.}
    \vspace{-7mm}
    \label{fig:additional_baselines}
\end{figure}

\begin{figure*}[ht]
    \centering
    \vspace{-3mm}
    \includegraphics[width=\textwidth]{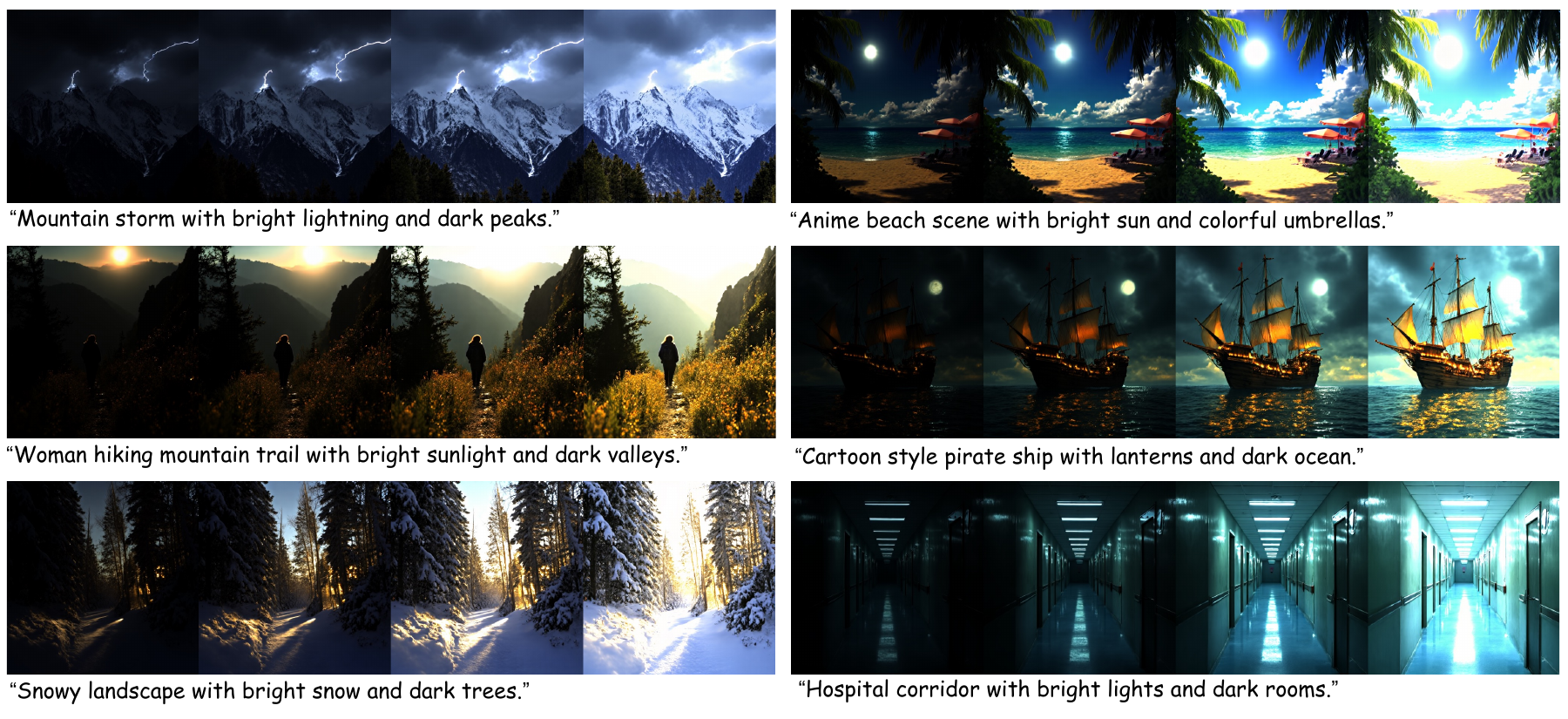}
    \vspace{-7mm}
    \caption{More visualization results generated by our method. Our approach supports the generation of images with various styles.}
    \vspace{-4mm}
    \label{fig:more_visualization_results}
\end{figure*}

\section{Experiments}
\label{sec:experiments}

\subsection{Training Details}
In our experiment, we use Flux-dev as our framework and conduct training on 4 NVIDIA A100 80GB GPUs. The batch size is 4 per GPU, and we train for 10,000 iterations in total. For LoRA finetuning, we set the rank to 64 and $\alpha$ to 128. In our Exposure Modulation Self-Attention module, we use 8 attention heads, an inner dimension of 512, and set the exposure modulation attention head dimension to 64. We adopt the AdamW optimizer, setting the learning rate for the exposure modulation module to $2\times10^{-5}$ and the learning rate for LoRA parameters to $1\times10^{-4}$. All training is performed using bf16 precision to improve computational efficiency and reduce memory consumption.

\subsection{Comparison with Previous Methods}
Existing works have not addressed the task of direct RAW or linear image generation, as discussed in Section~\ref{sec:related_work}. While some recent methods focus on HDR image generation, such as GlowGAN~\cite{wang2023glowgan}, their applicability is limited: GlowGAN can only generate HDR images for specific image categories and does not produce true linear images suitable for direct linear image workflows. Moreover, most prior HDR generation approaches operate in display-referred or tone-mapped spaces rather than predicting scene-referred linear content.
Due to the lack of direct competitors for linear image generation, we mainly compare our method with strong baselines by adapting existing architectures. Specifically, we conduct the following three types of comparisons:

\begin{enumerate}
    \item \textbf{T2I Finetuning:} We finetune the state-of-the-art text-to-image diffusion model Flux~\cite{flux2024} using LoRA on our dataset, training the model to generate linear images directly, rescaled to the $[0,1]$ range. This allows us to evaluate how a state-of-the-art T2I model adapts to the linear image domain.
    \item \textbf{T2V Finetuning:} We also test using text-to-video model Wan 2.1~\cite{wan2025wan}, whose VAE is capable of compressing $4T+1$ frames into $T$ latent representations. In our setting, we select the EV=0 frame as the reference and compress the 4 exposure brackets using Wan's VAE encoder into a single latent map. We then finetune a LoRA module to generate such latent maps, which are subsequently decoded back into exposure brackets. The same dataset as used for our baseline experiments is adopted for this finetuning procedure.
    \item \textbf{T2I Model Inflation:} We further compare against CameraCtrl~\cite{he2024cameractrl} and Generative Photography~\cite{tan2024generativephotography}, which inflate image diffusion architectures with temporal modules to generate multi-frame sequences. Since these methods are not designed for linear image generation, we fine-tune them on our training set for a fair comparison (denoted ``w/ F'').
\end{enumerate}

Table~\ref{tab:comparison_all} provides a quantitative comparison on Adobe FiveK~\cite{fivek} dataset, and Fig.~\ref{fig:comparison_methods} and Fig.~\ref{fig:additional_baselines} show qualitative results.
FID and LS are computed on linear images, while AS, NIQE, and CLIP Sim.\ are evaluated on the EV0 frame.
Due to the wide distribution of linear images, directly finetuning T2I Model on linear data makes it difficult to balance shadow and highlight details.
T2I Model Inflation methods suffer from both limited dynamic range and significant image quality degradation even after fine-tuning.
For T2V Finetuning, Wan 2.1's 4$\times$ temporal downsampling entangles the 4 exposure brackets into a single latent representation, causing a severe distribution mismatch that cannot be resolved through fine-tuning alone.
By directly modeling scene-referred properties using exposure brackets, our method achieves superior visual quality and dynamic range across all baselines.

\begin{table}[t]
    \centering
    \small
    \renewcommand{\arraystretch}{1.1}
    \resizebox{0.48\textwidth}{!}{%
    \begin{tabular}{llccccc}
        \toprule
        Type & Model & FID $\downarrow$ & AS $\uparrow$ & NIQE $\downarrow$ & CLIP Sim.$\uparrow$ & LS $\uparrow$ \\
        \midrule
        T2I Finetuning & Flux~\cite{flux2024}   & 32.12 & 4.712 & 5.304 & 25.90 & /    \\ 
        \midrule
        T2V Finetuning & Wan 2.1~\cite{wan2025wan}   & /     & 4.537 & 5.412 & 24.79 & 1.12 \\
        \midrule
        \multirow{4}{*}{T2I Model Inflation}
            & CameraCtrl~\cite{he2024cameractrl} (w/ F)                              & 37.25 & 5.230 & 4.131 & \textbf{26.89} &  8.97 \\
            & Gen.\ Photography~\cite{tan2024generativephotography} (w/ F)           & 40.17 & 4.619 & 4.514 & 23.71 &  7.11 \\
            & Gen.\ Photography~\cite{tan2024generativephotography} (w/o F)          & 43.83 & 3.909 & 4.870 & 20.51 &  5.56 \\
         &Ours & \textbf{28.29} & \textbf{5.700} & \textbf{3.658} & 26.02 & \textbf{23.06} \\
        \bottomrule
    \end{tabular}%
    }
    \vspace{-4mm}
    \caption{Comprehensive quantitative comparison with all baselines. T2I/T2V Finetuning directly adapt existing generative models to linear image generation via LoRA. T2I Model Inflation methods extend image diffusion architectures with temporal modules and are fine-tuned on our training set (w/ F). ``/'' indicates the metric cannot be meaningfully computed: Wan 2.1 cannot produce consistent exposure brackets required for HDR fusion (LS), and T2I Finetuning lacks the temporal capacity to span the full dynamic range (LS). ``F'' denotes fine-tuning on our training set. ``Gen.\ Ph.'' denotes Generative Photography~\cite{tan2024generativephotography}.}
    \vspace{-5mm}
    \label{tab:comparison_all}
\end{table}

\vspace{-2mm}
\subsection{Radiance Scale Estimation}
\vspace{-1mm}
Accurate radiance scale estimation requires integrating both semantic context from text tokens and spatial luminance cues from image tokens.
Our token denoising strategy achieves this naturally: the radiance scale token participates in the joint self-attention over all tokens during denoising, allowing it to simultaneously attend to text descriptions and image content.
To validate this design, we compare against a global pooling baseline, where pooled output tokens are passed through an MLP for radiance scale prediction. We test three variants: using only text tokens, only image tokens, or their concatenation.
As shown in Table~\ref{tab:luminance_estimation}, the token denoising method better leverages both text and image information simultaneously, leading to higher estimation accuracy than all global pooling alternatives.

\vspace{-1mm}
\subsection{More Analysis}
As shown in Fig.~\ref{fig:more_visualization_results}, our method can generate exposure brackets with diverse visual styles. More visualization results, ControlNet-based generation, linear image inpainting, and text-guided linear image editing are provided in the supplementary materials. Additional ablation studies covering positional encoding, number of exposure brackets, model architecture design, and EV injection strategy are also provided in the supplementary materials.
\vspace{-2mm}

\begin{table}[t]
    \centering
    \small
    \renewcommand{\arraystretch}{1.1}
    \renewcommand{\tabcolsep}{3.8mm}
    \resizebox{0.85\columnwidth}{!}{
    \begin{tabular}{lcccc}
        \toprule
        & Text-MLP & Image-MLP & Merge-MLP & Ours \\
        \midrule
        MAE $\downarrow$ & 0.782 & 1.213 & 0.792 & \textbf{0.737} \\
        \bottomrule
    \end{tabular}}
    \vspace{-1mm}
    \caption{Quantitative comparison of radiance scale estimation methods using MAE.}
    \vspace{-5mm}
    \label{tab:luminance_estimation}
\end{table}

%% file: sec/6_conclusion.tex
\section{Conclusion}
\label{sec:conclusion}
\vspace{-1mm}
In this paper, we present a novel generative framework for linear image synthesis with high dynamic range. Our method adopts a flow-matching paradigm with multi-exposure bracket prediction to overcome the limitations of VAE models in modeling scene radiance, enabling improved dynamic range and radiance fidelity in the generated images.
Building upon a DiT backbone, we introduce exposure modulation self-attention and radiance-scale token denoising, which ensure accurate generation across diverse exposure levels and enable explicit prediction of scene radiance scale. Our approach achieves efficient adaptation on limited linear data while maintaining image quality. Additionally, the framework enables downstream applications including linear image editing and ControlNet-based conditional generation, bridging the gap between generation model and professional photography workflows.

%% file: sec/X_suppl.tex
\clearpage
\renewcommand{\thesection}{\Alph{section}}
\renewcommand{\thesubsection}{\Alph{section}.\arabic{subsection}}
\setcounter{section}{0}
\setcounter{page}{1}
\maketitlesupplementary

\section{Additional Ablation Studies}
\label{sec:additional_ablations}
This section presents additional ablation studies of our method, investigating the influence of positional encoding, the number of exposure brackets, key architectural components, and alternative exposure modulation choices on generation quality and multi-exposure consistency. We provide both quantitative results and qualitative analyses to further validate the effectiveness and necessity of our proposed modules.

\noindent{\bf Positional Encoding.}
As shown in Table~\ref{tab:ablation_position_encoding}, positional encoding plays a crucial role in separating frame brightness. Using 2D RoPE alone makes tokens at the same spatial location across different exposure brackets difficult to distinguish, leading to noticeable checkerboard-like artifacts. Introducing a Layer Embedding (LE) significantly improves performance, indicating that explicit bracket identity helps resolve token ambiguity. However, when 3D RoPE is used, adding LE offers no clear improvement. This suggests that 3D RoPE already encodes bracket identity effectively through positional embedding. Therefore, we adopt pure 3D RoPE as our default design.

\begin{table}[h]
    \centering
    \small
    \renewcommand{\arraystretch}{1.1}
    \resizebox{0.7\columnwidth}{!}{%
    \begin{tabular}{lccc}
        \toprule
        Method & AS $\uparrow$ & NIQE $\downarrow$ & LS $\uparrow$ \\
        \midrule
        2D RoPE          & 4.099          & 4.332          &  4.10 \\
        2D RoPE + LE     & \textbf{5.792}          & 3.853          &  6.21 \\
        3D RoPE + LE     & 5.695          & 3.721          & 21.07 \\
        3D RoPE (Ours)   & 5.700 & \textbf{3.658} & \textbf{23.06} \\
        \bottomrule
    \end{tabular}
    }
    \vspace{-1mm}
    \caption{Ablation study on positional encoding methods. LE represents Layer Embedding. 3D RoPE achieves the best performance across all metrics. LS represents the luminance scale which measures the ratio between the brightest and darkest images.}
    \vspace{-2mm}
    \label{tab:ablation_position_encoding}
\end{table}

\begin{figure}[t]
  \centering
  \includegraphics[width=1.0\linewidth]{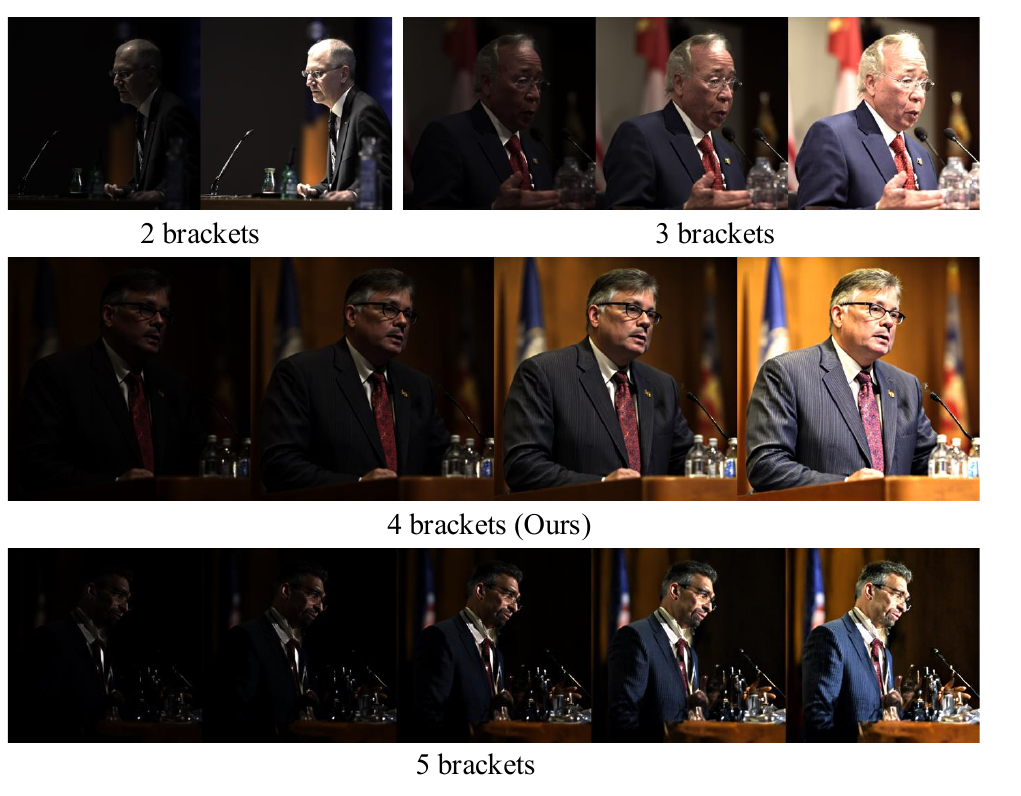}
  \vspace{-8mm}
    \caption{Visual comparison of models trained with different numbers of exposure brackets. Increasing the number of brackets expands the dynamic range but may degrade image quality. Training with more than five brackets tends to introduce noticeable artifacts; therefore, we adopt four brackets as our baseline, offering a balanced trade-off between dynamic range and visual fidelity.}
    \vspace{-2mm}
  \label{fig:num_bracs}
\end{figure}

\noindent{\bf{Number of Exposure Brackets}.}
To determine the optimal number of exposure brackets, we test how different bracket number influences image generation quality and the luminance scale (LS) between the brightest and darkest frames.
For all configurations, we set the exposure value (EV) interval between each adjacent frame to 2. Specifically, for 2 brackets we use EVs of $[-2,\,0]$; for 3 brackets: $[-2,\,0,\,2]$; for 4 brackets: $[-4,\,-2,\,0,\,2]$; and for 5 brackets: $[-6,\,-4,\,-2,\,0,\,2]$.
As shown in Table~\ref{tab:num_brackets}, using 4 exposure brackets achieves the best overall performance. Although 5 brackets can slightly improve luminance scale, increasing the number of generated brackets in fact leads to a degradation in image quality, as shown in Fig.~\ref{fig:num_bracs}.

\begin{table}[h]
    \centering
    \small
    \resizebox{0.8\columnwidth}{!}{%
    \begin{tabular}{lcccc}
        \toprule
        \# Brackets & AS $\uparrow$ & NIQE $\downarrow$ & LS $\uparrow$ & CLIP Sim.$\uparrow$ \\
            \midrule
            2 & \textbf{5.820} & 3.864 & 4.76 & 25.68 \\
            3 & 5.543 & 4.000 & 7.00 & 25.62 \\
            4 (Ours) & 5.700 & \textbf{3.658} & 23.06 & \textbf{26.02} \\
            5 & 5.258 & 4.294 & \textbf{25.76} & 23.92 \\
        \bottomrule
    \end{tabular}
    }
    \vspace{-1mm}
    \caption{Quantitative comparisons on the number of exposure brackets. 4 brackets achieve the best trade-off between image quality and dynamic range.}
    \vspace{-2mm}
    \label{tab:num_brackets}
\end{table}

\begin{figure}[t]
  \centering
  \includegraphics[width=1.0\linewidth]{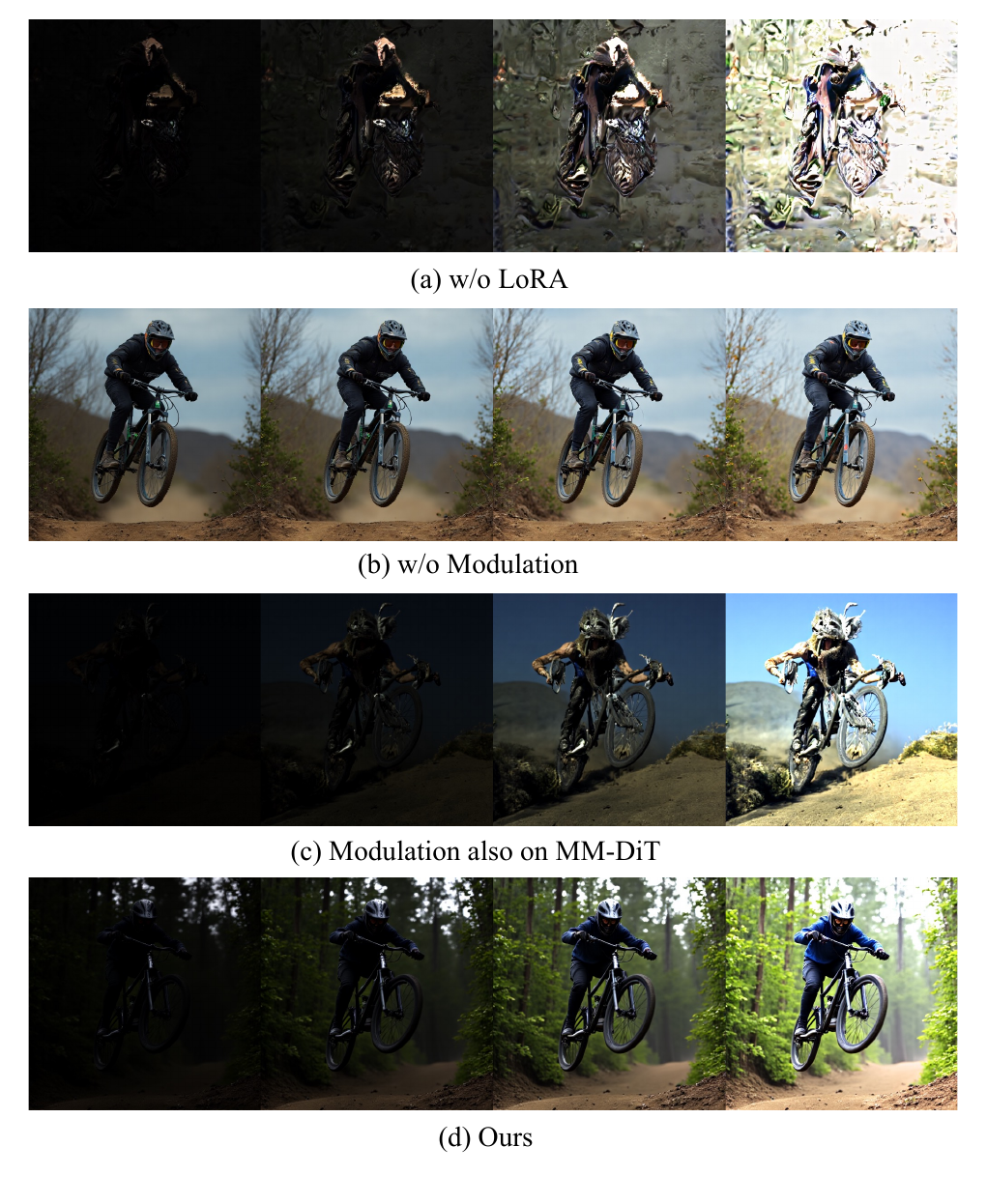}
  \vspace{-12mm}
    \caption{Visual comparison of generation results under different ablation settings. Without LoRA (a), the model becomes unstable during training due to the large distribution shifts across exposure levels in the sequence, leading to degraded and inconsistent outputs. Without modulation (b), the model produces visually plausible images but lacks meaningful exposure control. Applying modulation only to the MM-DiT branch (c) enhances exposure variation but introduces structural distortions. Our full method (d) achieves stable training, preserves structural fidelity, and produces coherent and physically plausible exposure transitions.}
    \vspace{-2mm}
  \label{fig:supp_ablation}
\end{figure}

\noindent{\bf{Model Architecture Ablation}.}
We evaluate different model architecture design choices including the impact of LoRA~\cite{hu2022lora} module and exposure modulation components. Table~\ref{tab:model_design} shows that both LoRA and our exposure modulation module are essential for optimal performance.
As shown in Fig.~\ref{fig:supp_ablation}, removing LoRA results in severe distorted global structure; removing the exposure modulation module causes poor contrast across brackets.
Since the MM-DiT~\cite{esser2024scaling} branch at the early stage of Flux~\cite{flux2024} is mainly responsible for modeling the overall structure of the image, introducing exposure modulation module at this stage destabilizes training and leads to structural distortions. 
Therefore, we only apply our exposure modulation module to the single-DiT component, which enables the model to maintain consistent global structure across all brackets, while allowing fine-grained control of luminance and detail alignment for each bracket.

\begin{table}[h]
    \centering
    \small
    \resizebox{0.97\columnwidth}{!}{%
    \begin{tabular}{lcccc}
        \toprule
        Method & AS $\uparrow$ & NIQE $\downarrow$ & LS $\uparrow$ & CLIP Sim.$\uparrow$ \\
            \midrule
            w/o LoRA & 3.183 & 3.899 & 52.58 & 17.88 \\
            w/o Modulation & \textbf{6.245} & 4.143 & 2.13 & 25.54 \\
            Modulation on MM-DiT & 4.937 & 3.987 & \textbf{86.61} & 24.15 \\
            Ours & 5.700 & \textbf{3.658} & 23.06 & \textbf{26.02} \\     
        \bottomrule
    \end{tabular}
    }
    \vspace{-1mm}
    \caption{Quantitative comparisons on model architecture ablation. Our exposure modulation module, when applied specifically to the single-DiT architecture, is crucial for stable multi-exposure generation and coherent exposure transitions across brackets.}
    \vspace{-2mm}
    \label{tab:model_design}
\end{table}

\begin{figure}[t]
  \centering
  \includegraphics[width=1.0\linewidth]{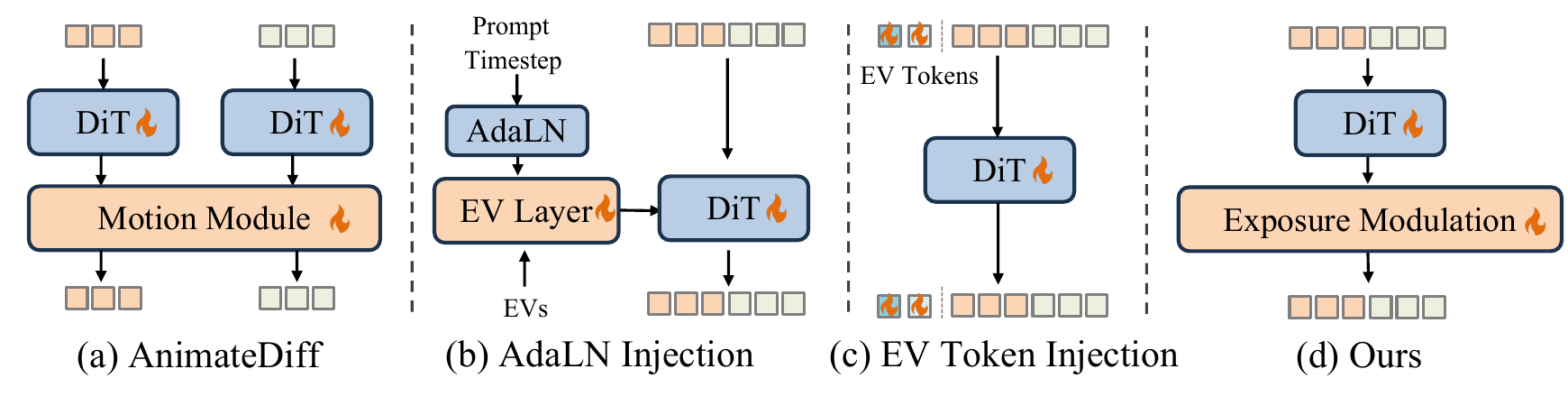}
  \vspace{-8mm}
\caption{Comparison of different EV injection strategies for exposure modulation. In (a) AnimateDiff~\cite{guo2024animatediff}, exposure information is implicitly encoded by concatenating brackets along the batch dimension, followed by sequence alignment using the motion module, but without explicit exposure conditioning. In (b) AdaLN~\cite{peebles2023scalable} Injection, EV guidance is introduced by appending a modulation layer after each LayerNorm in DiT, which adjusts the scaling and shifting parameters $(\alpha, \beta, \gamma)$ in an exposure-aware manner. In (c) EV Token Injection, learnable EV tokens are inserted into the token sequence, and an attention mask constrains each bracket to attend only to its corresponding EV token. In (d) Ours, we propose an exposure modulation module that directly adjusts feature activations based on EVs, enabling stable training and fine-grained, spatially coherent exposure control across brackets.}
    \vspace{-2mm}
  \label{fig:ev_inject_method}
\end{figure}

\begin{figure}[h]
    \centering
    \vspace{-3mm}
    \includegraphics[width=\columnwidth]{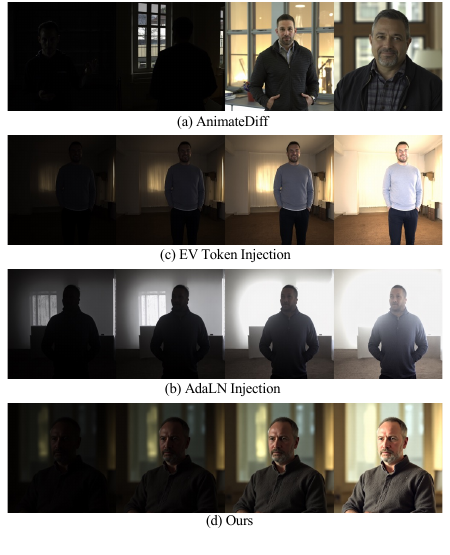}
    \vspace{-12mm}
    \caption{Visual comparison of various exposure modulation methods. Unlike other approaches, our method produces consistently aligned content across different exposure levels while preserving high image quality.}
    \label{fig:ev_injection}
\end{figure}

\noindent{\bf{Exposure Modulation Methods Comparison}.}
To enable effective exposure modulation, we systematically explore three major approaches: (1) considering exposure value as conditions like text prompt or time. (2) injecting explicit exposure tokens to the sequence, and (3) introducing additional dedicated network modules for modulation as shown in Fig.~\ref{fig:ev_inject_method}. To ensure stable training and robust adaptation across different methods, we consistently add a LoRA module with rank 64 on the main DiT backbone in all experiments. 
We provide a comprehensive evaluation of these exposure modulation strategies in Table~\ref{tab:ev_injection} and Fig.~\ref{fig:ev_injection}. Specifically, the AnimateDiff-style approach (a) encodes exposure information implicitly by concatenating exposure brackets along the batch dimension, relying on the motion module for sequence alignment.
The AdaLN-based method (b) inserts a zero-initialized modulation layer after each LayerNorm~\cite{ba2016layer} in the DiT blocks. 
This EV modulation layer takes the scalar exposure value (EV) as input, encodes it via Fourier features followed by a lightweight MLP, and produces bracket-specific FiLM~\cite{perez2018film} parameters $(\Delta\gamma, \Delta\beta)$ as well as gate offsets $(\Delta\text{gate})$. 
These parameters are then applied to the normalized hidden features and gating signals in a residual manner, enabling exposure-dependent adjustment. 
All EV-dependent offsets are zero-initialized and magnitude-constrained, ensuring that the model initially behaves identically to the original DiT and gradually learns stable exposure-aware modulation during training.
The EV Token Injection approach (c) introduces learnable exposure tokens and leveraging attention masks, but can destabilize image structure (see Fig.~\ref{fig:ev_inject_method}(c) and Table~\ref{tab:ev_injection}). 
In contrast, our proposed exposure modulation self-attention module (d) directly modulates features according to exposure values, producing stable, spatially consistent, and physically plausible exposure transitions, consistently achieving the best results in both exposure alignment and image quality.

\begin{table}[t]
    \centering
    \small
    \resizebox{0.85\columnwidth}{!}{%
    \begin{tabular}{lcccc}
        \toprule
        Method & AS $\uparrow$ & NIQE $\downarrow$ & LS $\uparrow$ & CLIP Sim.$\uparrow$ \\
            \midrule
            AnimateDiff & 4.521 & 6.201 & 2.34 & 23.94 \\
            AdaLN Injection & 4.481 & 5.098 & 14.86 & 21.15 \\
            EV Token & 5.372 & 4.291 & 18.04 & 24.23 \\
            \textbf{Ours} & \textbf{5.700} & \textbf{3.658} & \textbf{23.06} & \textbf{26.02} \\
        \bottomrule
    \end{tabular}
    }
    \vspace{-1mm}
    \caption{Detailed comparison of EV injection methods. Our exposure modulation self-attention mechanism demonstrates superior performance in maintaining exposure consistency while preserving image quality.}
    \vspace{-2mm}
    \label{tab:ev_injection}
\end{table}

\begin{figure*}[t]
    \centering
    \includegraphics[width=1.0\textwidth]{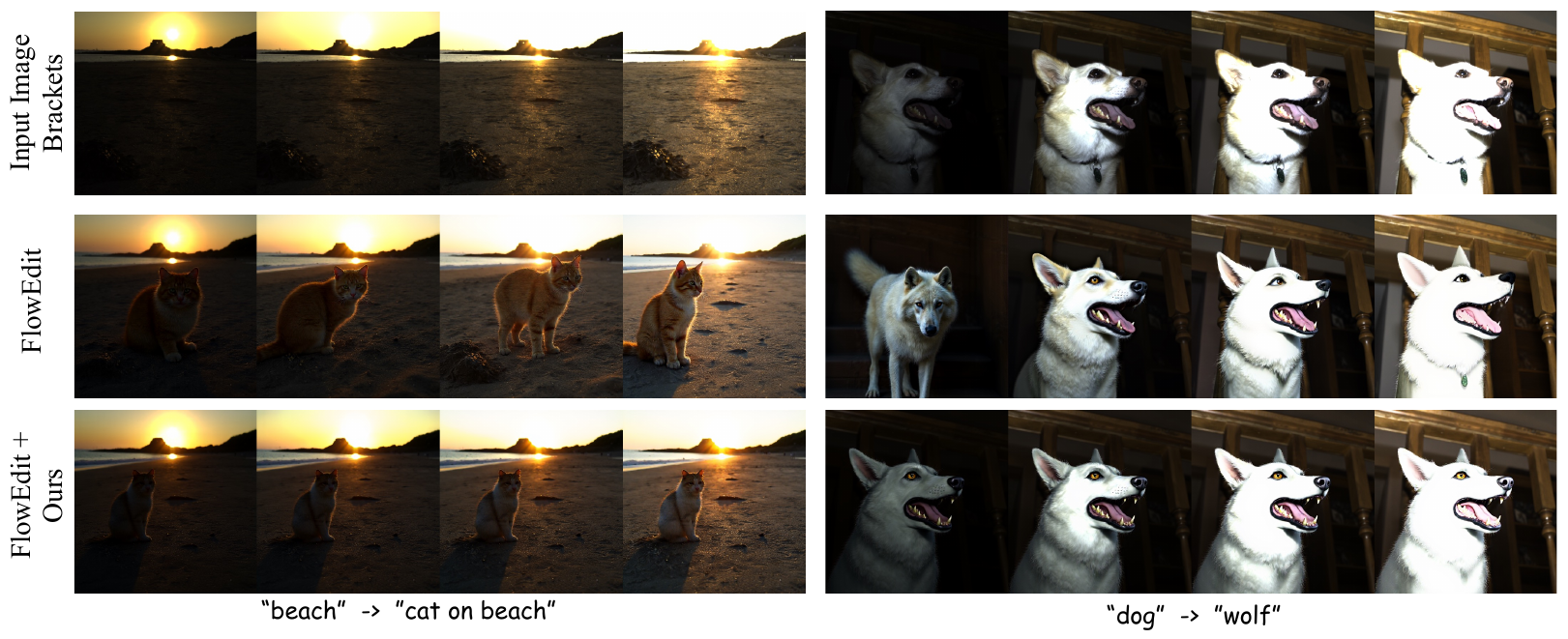}
    \vspace{-7mm}
    \caption{Visualization results of our method with FlowEdit~\cite{kulikov2025flowedit}. By integrating FlowEdit, our method enables intuitive and consistent editing across different exposure brackets without finetuning. Without our method, FlowEdit struggles to achieve consistent edits for the various exposure brackets, making it unsuitable for linear image editing.}
    \vspace{-2mm}
    \label{fig:linear_editing}
\end{figure*}

\begin{figure}[ht]
    \centering
    \vspace{-3mm}
    \includegraphics[width=\columnwidth]{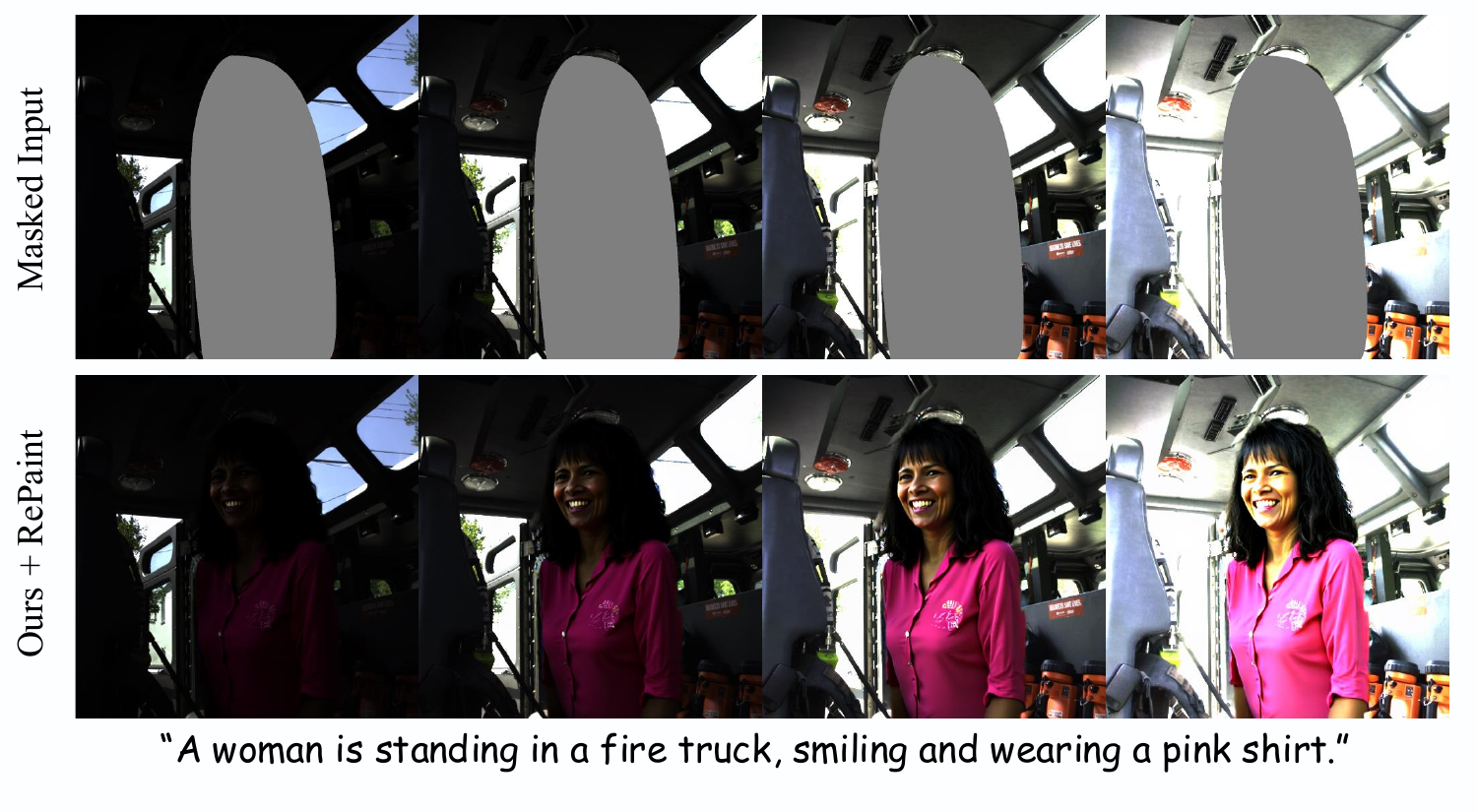}
    \caption{Linear image inpainting results. By combining our exposure-aware generation model with RePaint, missing regions across exposure brackets can be faithfully reconstructed. The inpainted content is well-aligned with the surrounding structures and maintains consistent exposure relationships across both shadow and highlight areas.}
    \label{fig:inpainting_results}
\end{figure}

\begin{figure*}[t]
    \centering
    \vspace{-3mm}
    \includegraphics[width=\textwidth]{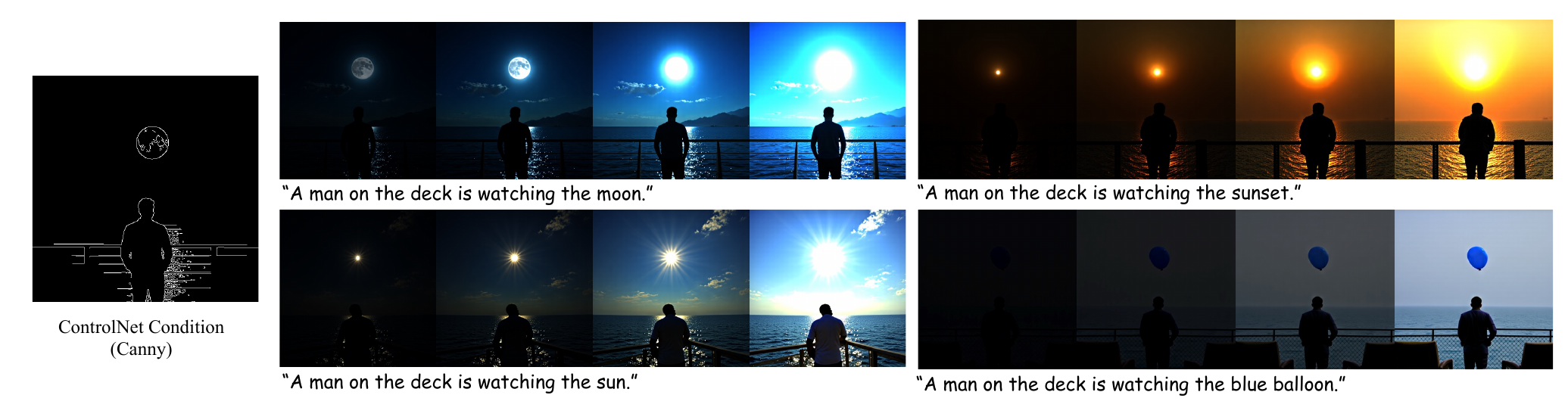}
    \vspace{-9mm}
    \caption{ControlNet-based linear image generation results with Canny edge guidance. Our method enables the synthesis of diverse exposure brackets with varied content while preserving consistency with the given edge condition, demonstrating strong controllability and flexibility in content generation.}
    \vspace{-1mm}
    \label{fig:controlnet_results}
\end{figure*}

\begin{figure*}[t]
    \centering
    \vspace{-3mm}
    \includegraphics[width=\textwidth]{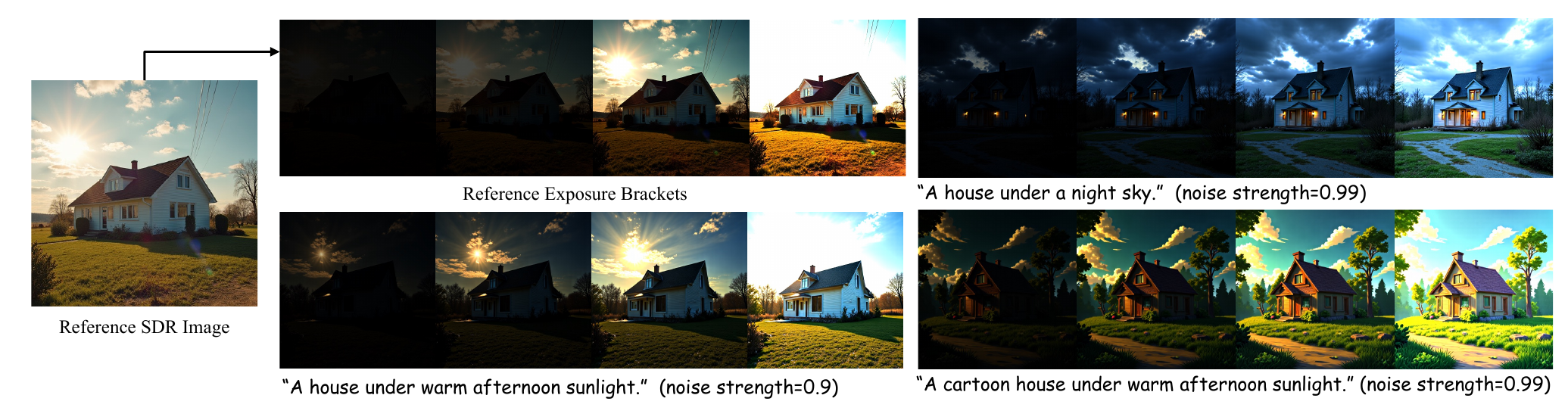}
    \vspace{-9mm}
    \caption{Reference-based HDR exposure brackets generation from an SDR image. Starting from a display-referred SDR image (left), we apply a gamma correction to approximate its linear domain representation and synthesize a set of pseudo-linear reference exposure brackets (middle-top). Although these brackets have limited dynamic range, they serve as structural and exposure-aware guidance. Using an SDEdit~\cite{sdedit} noise–denoise process, we progressively inject noise into the reference brackets and perform conditional denoising to generate HDR exposure brackets with enriched dynamic range (middle-bottom). By adjusting the noise strength during SDEdit, we can control the trade-off between fidelity to the reference exposure structure and the generation of plausible high-dynamic-range content (right). This enables reference-guided HDR bracket generation while preserving semantic consistency.}
    \vspace{-1mm}
    \label{fig:sdr2hdr}
\end{figure*}

\section{Post-Editing Flexibility}
\label{sec:post_editing}
We compare the post-editing capability of our method against HDR generation methods SingleHDR~\cite{liu2020single} and LEDiff~\cite{wang2025lediff} in display space to quantify the flexibility of the generated content.
All methods are tone-mapped to display space using a global Reinhard operator with $\gamma = 2.2$.
Since SingleHDR and LEDiff follow an image-to-HDR paradigm (requiring an SDR input), we use our EV0 bracket output as their input to ensure aligned comparison.
We then apply brightness adjustments ($\pm2$ EV) and white balance shifts ($\pm2000$ K) to each method's output, reporting FID against unedited ground-truth images.

As shown in Table~\ref{tab:post_editing}, our method outperforms prior HDR generation methods on all post-editing metrics, demonstrating the advantage of generating scene-referred linear images for downstream editing.
``Ours (SDR)'' denotes our EV0 frame rendered directly without HDR fusion; its degraded performance after large edits confirms that full bracket generation and fusion is essential for robust post-editing.

\begin{table}[h]
    \centering
    \small
    \resizebox{\columnwidth}{!}{%
    \begin{tabular}{lcccccc}
        \toprule
        Method & AS $\uparrow$ & FID $\downarrow$ & FID +2EV $\downarrow$ & FID -2EV $\downarrow$ & FID +2000K $\downarrow$ & FID -2000K $\downarrow$ \\
        \midrule
        SingleHDR~\cite{liu2020single} & 5.781 & 31.57 & 29.17 & 32.84 & 31.82 & 30.78 \\
        LEDiff~\cite{wang2025lediff}   & 5.806 & 29.94 & 28.67 & 28.81 & 30.34 & 28.98 \\
        Ours (SDR)                     & 5.700 & 32.26 & 31.14 & 38.10 & 32.15 & 32.31 \\
        Ours                           & \textbf{5.819} & \textbf{27.87} & \textbf{26.40} & \textbf{28.19} & \textbf{27.96} & \textbf{28.28} \\
        \bottomrule
    \end{tabular}
    }
    \vspace{-1mm}
    \caption{Post-editing comparison with HDR generation methods in display space (after Reinhard tone-mapping with $\gamma=2.2$). FID is reported before and after $\pm2$ EV brightness and $\pm2000$ K white balance adjustments. Our method consistently achieves lower FID after editing, demonstrating the advantages of linear-space generation for post-processing.}
    \vspace{-2mm}
    \label{tab:post_editing}
\end{table}

\section{Downstream Applications}
\label{sec:applications}

\noindent{\bf{Linear Image Inpainting}.}
In addition to exposure bracket generation, our model can be directly applied to linear image inpainting. By leveraging the exposure-aware representation, the model is able to restore masked regions in each bracket while preserving the relative exposure relationships across brackets. As shown in Fig.~\ref{fig:inpainting_results}, combining our model with RePaint~\cite{lugmayr2022repaint} enables structurally coherent and exposure-consistent inpainting across a wide dynamic range. The reconstructed regions seamlessly blend with the original content under both underexposed and overexposed conditions, demonstrating the applicability of our approach to linear and HDR inpainting tasks.

\noindent{\bf{Linear Image Editing}.}
Our method can also be used to edit linear images using a training-free flow-matching image editing method.
We use FlowEdit~\cite{kulikov2025flowedit} to edit linear images by first decomposing them into multiple exposure brackets, applying aligned edits on each bracket using our pretrained model, and then fusing them back with our multiple exposure brackets fusion strategy.
This approach ensures edits are consistent and radiometrically correct across exposures.
As shown in Fig.~\ref{fig:linear_editing}, by integrating FlowEdit, our method enables intuitive and consistent editing across different exposure brackets without finetuning. Without our method, FlowEdit struggles to achieve consistent edits across the various exposure brackets, making it unsuitable for linear image editing.

\noindent{\bf{ControlNet-based Linear Image Generation}.}
We demonstrate the compatibility of our method with ControlNet~\cite{zhang2023adding} guidance for conditional linear image generation. 
Fig.~\ref{fig:controlnet_results} shows results with different control conditions.
By controlling the exposure brackets, we can ultimately synthesize a linear image that adheres to the given condition.
Since Flux's ControlNet is primarily designed for 1024$\times$1024 resolution, occasional misalignment may still occur at lower resolutions.

\noindent{\bf{HDR Rendering}.}
After generating a set of exposure brackets with our method, we can synthesize a linear image by merging these brackets in linear space. To produce a display-ready HDR image, we apply gamma correction to the merged linear result, effectively mapping it into the appropriate non-linear space for visualization or HDR export. This workflow allows for the preservation of high dynamic range and accurate scene representation in the final HDR output.
We have attached a website containing HDR images rendered from our generated linear images.

\noindent{\bf{Reference-based HDR Image Generation}.}
As illustrated in Fig.~\ref{fig:sdr2hdr}, our approach also enables reference-based HDR image generation from standard SDR images. We simulate pseudo-linear exposure brackets from an input SDR image, then use an SDEdit-style strategy: noise is injected into these brackets, and our model leverages its generative prior to reconstruct the noisy latents back into the linear domain. The final outputs can then be merged and mapped to produce an HDR image, enriching the dynamic range while preserving semantic consistency with the original SDR reference.

\section{Additional Qualitative Results}
\label{sec:additional_results}

\begin{figure*}[ht]
    \centering
    \vspace{-3mm}
    \includegraphics[width=\textwidth]{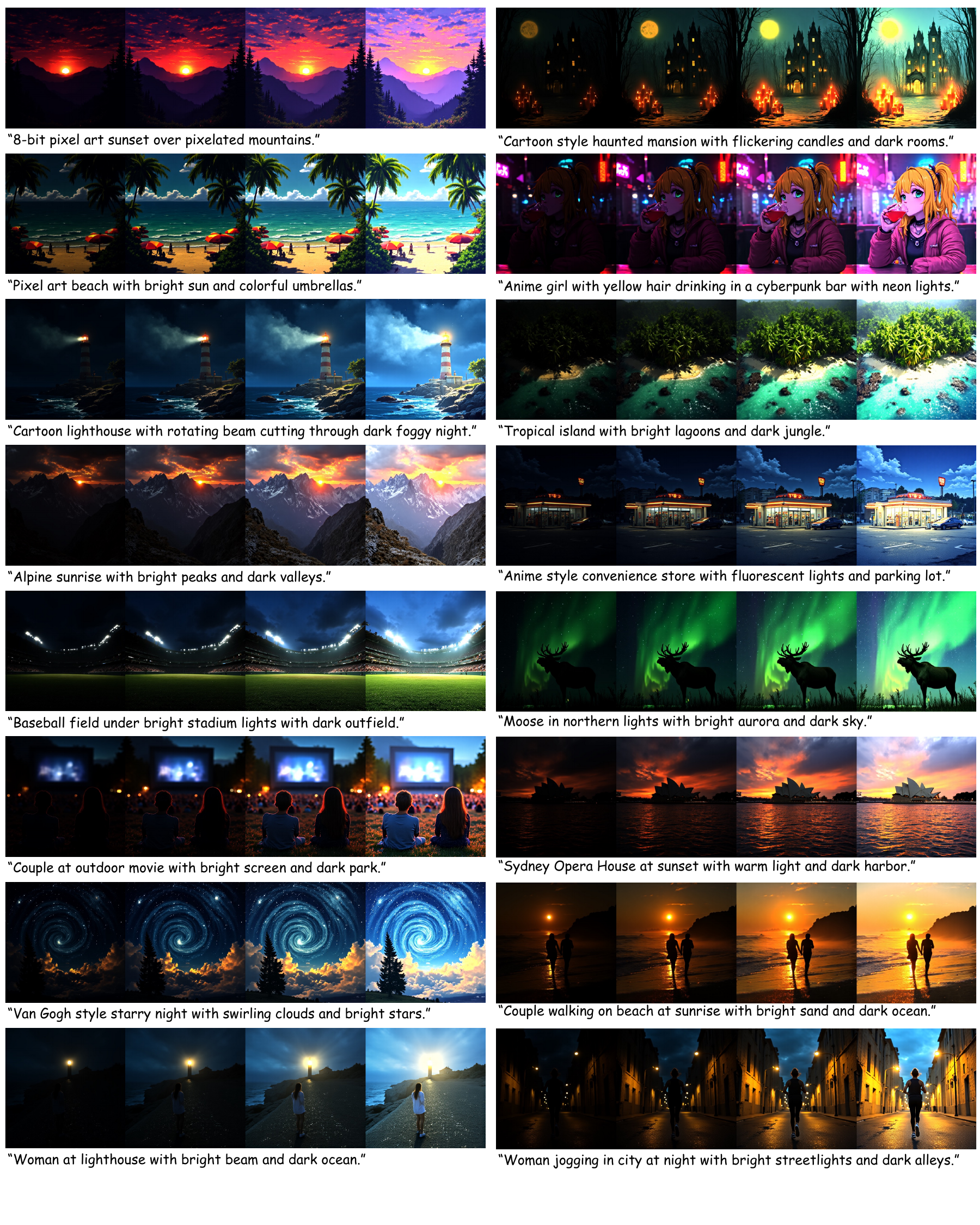}
    \vspace{-14mm}
    \caption{Diverse style generation results. Our method can generate exposure brackets with various artistic styles including cartoon, photorealistic, painterly, and cinematic looks while preserving proper exposure relationships across brackets.}
    \label{fig:diverse_styles}
\end{figure*}

Fig.~\ref{fig:diverse_styles} demonstrates our method's capability to generate linear images across various artistic styles while maintaining proper exposure relationships across brackets.

\section{Limitations}
\label{sec:limitations}
While our method achieves promising results for linear image generation, several limitations remain:
Since Flux itself cannot guarantee high-quality generation for high-resolution images (e.g., 2K), increasing the number of exposure brackets further affects image quality. When the frame number becomes too large, it becomes difficult to maintain stable generation even at 1024 resolution, leading to inconsistent exposure relationships and potential artifacts across brackets.
Besides, our training dataset has inherent limitations in terms of aesthetic scores, which are not particularly high, and consists entirely of realistic photographic scenes. Consequently, when generating non-realistic objects such as robots, mechanical structures, or highly stylized content, the model's performance will be affected. 